\documentclass[runningheads]{llncs}

\usepackage{graphicx}
\usepackage{amsmath,amssymb}
\usepackage{mathtools}
\usepackage{xspace}
\usepackage{booktabs} 
\usepackage{xcolor}
\usepackage{rotating}
\usepackage{colortbl}
\usepackage{multirow}
\usepackage{array}
\usepackage{hyperref}
\usepackage{url}
\usepackage{comment}
\usepackage{arydshln}
\renewcommand{\epsilon}{\varepsilon}

\renewcommand{\phi}{\varphi}

\usepackage{fullpage}

\title{Towards Feature-Based Performance Regression\\ Using Trajectory Data}

\author{
Anja Jankovic\inst{1} \and Tome Eftimov\inst{2} \and 
Carola Doerr\inst{1}
}
\authorrunning{A. Jankovic, T. Eftimov, C. Doerr}
\institute{
$^1$Sorbonne Universit\'e, CNRS, LIP6, Paris, France\\
$^2$Computer Systems Department, Jo\v{z}ef Stefan Institute, Ljubljana, Slovenia
}

\begin{document}

\maketitle

\begin{abstract}
Black-box optimization is a very active area of research, with many new algorithms being developed every year. This variety is needed, on the one hand, since different algorithms are most suitable for different types of optimization problems. But the variety also poses a meta-problem: which algorithm to choose for a given problem at hand? Past research has shown that per-instance algorithm selection based on exploratory landscape analysis (ELA) can be an efficient mean to tackle this meta-problem. Existing approaches, however, require the approximation of problem features based on a significant number of samples, which are typically selected through uniform sampling or Latin Hypercube Designs. The evaluation of these points is costly, and the benefit of an ELA-based algorithm selection over a default algorithm must therefore be significant in order to pay off. One could hope to by-pass the evaluations for the feature approximations by using the samples that a default algorithm would anyway perform, i.e., by using the points of the default algorithm's trajectory. We analyze in this paper how well such an approach can work. Concretely, we test how accurately trajectory-based ELA approaches can predict the final solution quality of the CMA-ES after a fixed budget of function evaluations. 
We observe that the loss of trajectory-based predictions can be surprisingly small compared to the classical global sampling approach, if the remaining budget for which solution quality shall be predicted is not too large. Feature selection, in contrast, did not show any advantage in our experiments and rather led to worsened prediction accuracy. The inclusion of state variables of CMA-ES only has a moderate effect on the prediction accuracy.  

\end{abstract}

\sloppy{
\section{Introduction} 
\label{sec:intro}

In many real-world optimization challenges, we encounter optimization problems which are too complex to be explicitly modeled via mathematical functions, but which nonetheless need to be assessed and solved, more often than not requiring significant computational resources to do so. Explicit problem modeling is also an issue when the relationship between decision variables and solution quality cannot be established other than by simulations or experiments. A standard example for the latter is the design of (deep) neural networks. \emph{Black-box optimization algorithms (BBOA)} are algorithms designed to solve problems of the two types above. BBOA are usually iterative procedures, which actively steer the search by using information obtained from previous iterations, with the goal to eventually converge towards an estimated optimal solution. In each generation, a number of solutions candidates are generated and undergo evaluation. 

Classically, BBOA were manually designed, based on users' experience. A plethora of algorithmic components exist from which users can choose to build their own algorithms, and the number of these components is growing every year. 
Even though the basic underlying principles of these components can be considered similar in nature, their performances on different problem instances can greatly vary. 
An important and challenging task is thus to select the most appropriate and efficient algorithm when presented with a new, unknown problem instance.  This research problem, formalized as the \emph{algorithm selection problem (ASP)}~\cite{RiceASproblem}, is one of \emph{the} core questions that evolutionary computation aims to answer. 
The algorithm selection problem is classically tackled by relying on expert knowledge of both the problem instance and the algorithm's strengths and weaknesses.  In recent years, however, due to the significant progress in the machine learning (ML) field, there has been a shift towards an \emph{automated} selection~\cite{KerschkeKBHT18,KerschkeT19,MunozASSurvey,JD20} and configuration~\cite{BelkhirDSS17} of algorithms based on supervised learning approaches. The idea behind these approaches is in utilizing ML techniques to design and to train models to accurately predict the performance of different black-box algorithms on previously unseen problem instances, with the goal to use these performance predictions to select and to configure the best algorithm for the problem at hand, respectively.  
In order to apply supervised learning, problem instances need to be represented in a convenient way via numerical values. That is, we need to quantify relevant characteristics of a problem instance through appropriate measures. These measures are referred to as \emph{features}. In the terminology used in evolutionary computation (EC), features are hence aimed at describing the \emph{fitness landscape} of a problem instance. 

Fitness landscape analysis has a long tradition in EC. For practical use in black-box optimization, however, the fitness landscape properties can only be described via an informed guessing strategy. Concretely, we can only approximate the fitness landscapes, through the samples that we have evaluated and to which a solution quality has been assigned. Research addressing efficient ways to characterize problem instances via feature approximations is subsumed under the umbrella term \emph{exploratory landscape analysis (ELA)}~\cite{mersmann_exploratory_2011}. Common research questions in ELA concern the number of samples needed to accurately approximate feature values, the design and the selection of features that are descriptive and easy to approximate, and the possibility to use feature values to transfer learned policies from some instances to previously unseen ones.  

A drawback of existing ELA-approaches are the resources needed to extract and to compute the feature values, the time required to train the models, and a lack of explainability. In this work, we focus on the first issues, the feature extraction. Most ELA-based studies perform a three-step selection/configuration: in the first step a number of search points are sampled (commonly using uniform sampling, Latin Hypercube Designs, or quasi-random sampling~\cite{RenauDDD20}), evaluated, and then plugged into a feature computation algorithm such as the R tool \emph{flacco}~\cite{flacco}. In the second step the model for the classification or regression task is built and an algorithm and/or its configuration is suggested. In the third step, this algorithm is then run on the problem instance under consideration. Clearly, the effort for steps 1 and 2 cannot be neglected, and can have a decisive influence on the usefulness of a per-instance algorithm selection/configuration approach, as its effort needs to pay off compared to the performance of a default solver. Even when neglecting the computational overhead of this approach and focusing on function evaluations only as performance measure (as is commonly done in evolutionary computation~\cite{cocoplat}), the evaluations needed for completing step 1 need to be taken into account. Much research has been done on determining a suitable number of samples, and typical recommendations vary between $30d$~\cite{BelkhirDSS16} and $50d$ samples~\cite{kerschke_low-budget_2016}, where $d$ denotes the dimension of the problem. This is hence a considerable investment. 

Of course, one could use these samples to \emph{warm-start} the optimization heuristics, e.g., by initiating them in good regions and/or by calibrating their search behavior based on the information obtained from the samples used to compute the features. 

A charming, yet straightforward alternative would be to integrate the first step of the ELA-based approach described above into the optimization routine, by computing the features based on the search points that a default algorithm would anyway perform. That is, one would use the search trajectory of such a default algorithm to predict and then to select and/or to configure a solver \emph{on the fly}, once or even several times during the optimization process.  

Similar to parameter control~\cite{KarafotiasHE15,AletiM16,DoerrD20paramControlchapter}, such a dynamic selection would not only allow to \emph{identify} an efficient algorithm for the given problem instance, but could also benefit from \emph{tracking} the best choice while the optimization process (and the best response to its needs) evolves. Such a dynamic algorithm selection can therefore be seen as an ELA-based variant of hyper-heuristics~\cite{BurkeGHKOOQ13hyperheuristicssurvey}. The approach has previously been used in the context of constrained optimization, with the goal to have a dynamic, ELA-guided selection of a suitable constraint handling technique; see~\cite{Malan18DEconstraints} for examples and further references. 

A key challenge in applying dynamic ELA-based algorithm selection is the fact that the feature values can vary drastically between different sampling strategies~\cite{RenauDDD20}. Since the distribution of points sampled on different problem instances can differ quite drastically even when using the same algorithm, it is not clear, a priori, if or how suitable ML models can be trained. This challenge was confirmed in~\cite{JD19}, where it was shown that the landscape which an algorithm sees locally during the optimization process (i.e., the partial landscape it is aware of at each step of the optimization process) usually differs a lot from the global fitness landscape of the problem it is solving. 

\paragraph{Our Results.} With the long-term goal to obtain well-performing dynamic ELA-based algorithm selection and configuration techniques, we analyze in this work a first, rather cautious task: ELA-based performance prediction using the trajectory samples of the algorithm under investigation. More precisely, we 
consider the Covariance Matrix Adaptation Evolution Strategy (CMA-ES~\cite{HansenO01CMAES}), and we aim at predicting its solution quality (measured as target precision, i.e., the difference to an optimal solution in quality space) after a fixed budget of function evaluations. Concretely, we use the first 250 samples evaluated by the CMA-ES and we aim at predicting its performance after additional 250 evaluations, doing so for 20 independent CMA-ES runs. The performance regression is done via a random forest model which takes as input the features computed from the trajectory data and which outputs an estimate for the final solution quality.  

We then take into account that problem characteristics cannot only be described via classic ELA features, but that internal states of the search heuristics can also be used to derive information about the problem instance at hand. Such approaches have in the past been used, for example, for local surrogate-modelling~\cite{Holena}.  We analyze the accuracy gains when using the same state information as in~\cite{Holena}, that is, the values of the CMA-ES internal variables that mainly carry information about the current probability distribution from which the CMA-ES samples candidates for the new generation.  In our experiments, the advantage of using this state information over using ELA-features only, however, is only marginal. Concretely, the average 
difference between true and predicted solution quality decreases from 14.4 to 12.1 when adding the state variables as features (where the average error 
reported here is taken over all 24 benchmark problems from the BBOB suite of the COCO platform~\cite{cocoplat}, and over all performed CMA-ES runs).

We observe in the experiments above that some CMA-ES runs are drastic outliers in terms of performance, at times with the target precision differing from the target precision of all the other runs by up to 10 orders of magnitude.  We therefore also consider an intentionally more ``friendly'' setting, in which we analyze the regression quality only for the run achieving median performance on a given problem instance. 
Conclusions for combining trajectory-based and state variable features remain almost identical to those stated above.

We then compare these median trajectory-based predictions to the classical approach using globally sampled features. Here, we pessimistically assume that the samples were computed for free. That is, we couple 2 separate sets of the global feature values approximated from 250 and 2000 uniformly sampled points each to the target precision achieved by the CMA-ES after 500 function evaluations. 
Interestingly, the difference in prediction accuracy compared to our trajectory-based predictions is rather small.  The global predictions still remain, however, more accurate, with an average absolute prediction error of 4.7 vs. 6.2 for the trajectory approach (where again the average is taken over all 24 BBOB functions).

Furthermore, we also use this median setting to analyze the influence of feature selection on prediction accuracy.  Different state-of-the-art methods were applied, using a transfer learning scenario, to select features estimated to be the most important and to have highest discriminative power. Here again, the differences in prediction accuracy were small,  with feature selection surprisingly leading to an overall slightly worsened solution quality than the full feature portfolio.

As suggested in~\cite{JD20}, all our experiments are based on two independently trained models: one which aims to predict target precision after 500 evaluations, and one which predicts the logarithm of this target precision. While the former is better in guessing the broader ``ball park'', the latter is more suitable for fine-grained performance prediction, i.e., when the expected performance of the algorithm is very good. As in~\cite{JD20}, we also build a combined regression, which uses either one of the two models, depending on whether the predicted performance is better or worse than a certain threshold. The optimal thresholds differ quite drastically between different feature sets. However, a sensitivity analysis reveals that their influence on overall performance is rather small. Also, the ranking of the different feature portfolios remains almost unaffected by the choice of the threshold.  
In line with the results in~\cite{JD20}, the combined models perform consistently better than any of the two standalone ones, albeit slightly.

\section{Supervised ML for Performance Regression}
\label{sec:supervisedml}

\textbf{The Experimental Setup.} 
When it comes to landscape-aware performance prediction, supervised machine learning techniques such as regression and classification have been studied in a variety of settings.  Regression models, unlike classification ones, have an advantage of keeping track of the magnitude of differences between performances of different algorithms, as they measure concrete values for performances of all algorithms from the portfolio.

Among different supervised learning regressors in the literature, such as support vector machines, Gaussian processes or ridge regression to name a few, it has been empirically shown that random forests outperform other models in terms of prediction accuracy~\cite{HutterXHL14}.  A random forest is an ensemble-based meta-estimator that works by fitting multiple decision trees on subsamples of the original data set, then uses averaging as a way to control overfitting.  In our experimental setup, we used an off-the-shelf random forest regressor from the Python \emph{scikit-learn} package~\cite{scikit-learn}, without parameter tuning and using 1000 estimators.

We restricted this work to a single heuristic, the Covariance Matrix Adaptation Evolution Strategy (CMA-ES~\cite{HansenO01CMAES}). The CMA-ES works by iteratively sampling a new population of candidate solutions from a shifted multivariate normal distribution, choosing the best offspring of the current population based on their respective fitness values, and then updating the parameters of the probability distribution according to the best candidates. For the purpose of our work, we used its standard version, available in the Python \emph{pycma} package~\cite{pycmaes}, which uses a fixed population size and no restarts during the optimization process.

As our benchmark, we used the first five instances of all 24 noiseless \emph{BBOB} functions of the \emph{COCO} platform~\cite{cocoplat}, an environment for comparison of algorithm performance in continuous black-box optimization.  The different instances of each function are generated by translating and rotating the function in the objective space. These transformations do not affect the performance of CMA-ES, but they do influence some of the feature values, especially those which are not transformation-invariant~\cite{TomeinvarianceASoC} (for the invariant features, the boundary handling can have an effect on the feature values). We focus on dimension $d=5$ here.  

For our first experiments, we perform 20 independent runs of the CMA-ES on these 120 problem instances, while keeping track of the search trajectories and the internal state variables of the algorithm itself.  Throughout this work, we fix the budget of 500 function evaluations, after which we stop the optimization and record the target precision of the best found solution within the budget.  In order to predict those recorded target precisions after 500 function evaluations, we compute the trajectory-based landscape features using the first 250 sampled points and their evaluations from the beginning of each trajectory, and couple them with the values of the internal CMA-ES state variables extracted at the $250^{th}$ function evaluation.

\begin{figure*}[t]
\includegraphics[width=\textwidth]{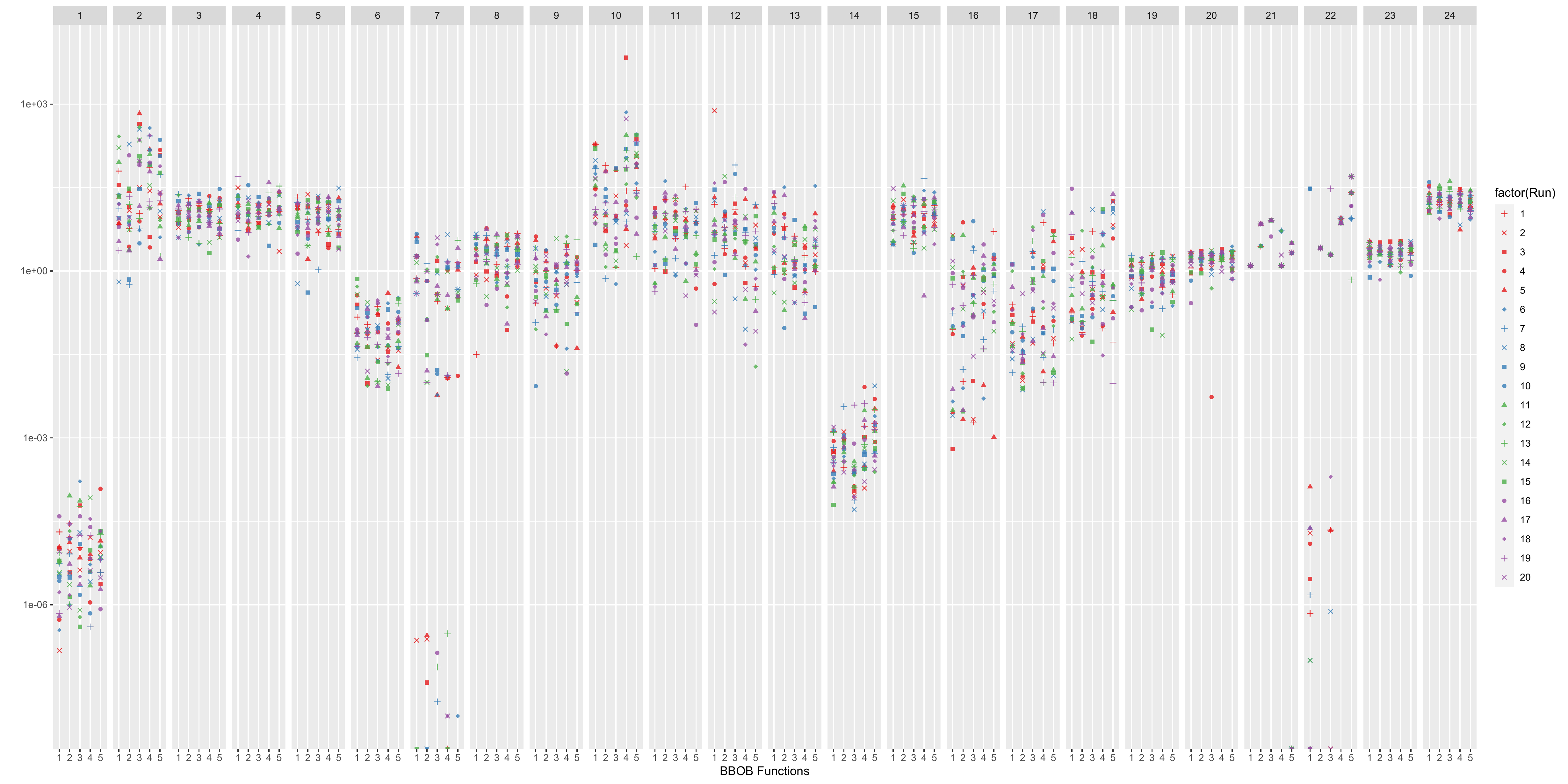}
\caption{Target precision achieved by the CMA-ES with a budget of 500 function evaluations, for each of the first five instances of all 24 BBOB functions. Differently colored and shaped points represent 20 independent CMA-ES runs.} \label{fig:true_tp_20_runs}
\end{figure*}

Figure~\ref{fig:true_tp_20_runs} summarizes the target precision achieved by CMA-ES in each of the 20 runs. We see that the results are more or less homogeneous across different runs and across different instances of the same problem. However, we also observe several outliers, e.g., for functions 7 (outlier for all instances), function 10 (instance 4), function 12 (instance 1). It is important to keep in mind that the randomness of these performances are entirely caused by the randomness of the algorithm itself -- the problem instance does not change between different runs. 

For landscape feature computation, we use the R package \emph{flacco}~\cite{flacco}. Following suggestions made in~\cite{KerschkeT19,BelkhirDSS17} we restrict ourselves to those feature sets that do not require additional function evaluations for computing the features.  Namely, in this work we use 2 original ELA feature sets (\emph{y-Distribution} and \emph{Meta-Model}), as well as \emph{Dispersion}, \emph{Nearest-Better Clustering} and \emph{Information Content} feature sets.  This gives us a total of 38 landscape features per problem instance.  In addition, we follow up on an idea previously used in~\cite{Holena} and consider a set of internal CMA-ES state variables as features:
\begin{itemize}
    \item Step-size: its value indicates how large is the approximated region from which the CMA-ES samples new candidate solutions.
    \item Mahalanobis mean distance: represents the measure of suitability of the current sampled population for model training from the point of view of the current state of the CMA-ES algorithm.
    \item $C$ evolution path length: indicates the similarity of landscapes among previous generations.
    \item $\sigma$ evolution path ratio: provides information about the changes in the probability distribution used to sample new candidates.
    \item CMA similarity likelihood: it is a log-likelihood of the set of candidate solutions with respect to the CMA-ES distribution and may also represent a measure of the set suitability for training.
\end{itemize}

As suggested in~\cite{JD20}, and using the elements described above, we establish two separate regression approaches.  One model is trained to predict the actual, true value of the target precision data (we refer to it as the \emph{unscaled model} in the remainder of the paper), while the other predicts the logarithm of the target precision data (the \emph{log-model}).  It is important to note that the target precision measure intuitively carries the information about the order of magnitude of the actual distance to the optimum, i.e., the \emph{distance level} to the optimum, which is effectively computed as the log-target precision.  For instance, if an algorithm reaches a target precision of $10^{-3}$ for one problem instance and $10^{-7}$ for another, it means that the algorithm found a solution which is 4 distance levels closer to the optimum in the latter scenario.  Moreover, to reduce variability, we estimate both models' prediction accuracy through performing a 5-fold \emph{leave-one-instance-out} cross-validation, making sure to train on 4 out of 5 instances per BBOB function, test on the remaining instance and combine the results over the rounds.

\textbf{Results.} 
Adopting our two regression models, we trained them separately in the following three scenarios: using as predictor variables the landscape features only, using the internal CMA-ES state variables only, and using the combination of the two.  We trained the random forests 3 independent times and took a median of the 3 runs to ensure the robustness of the results.

\begin{figure*}[t]
\centering 
    \includegraphics[trim= 1cm 12.5cm 6cm 1cm, clip, width=0.6\textwidth]{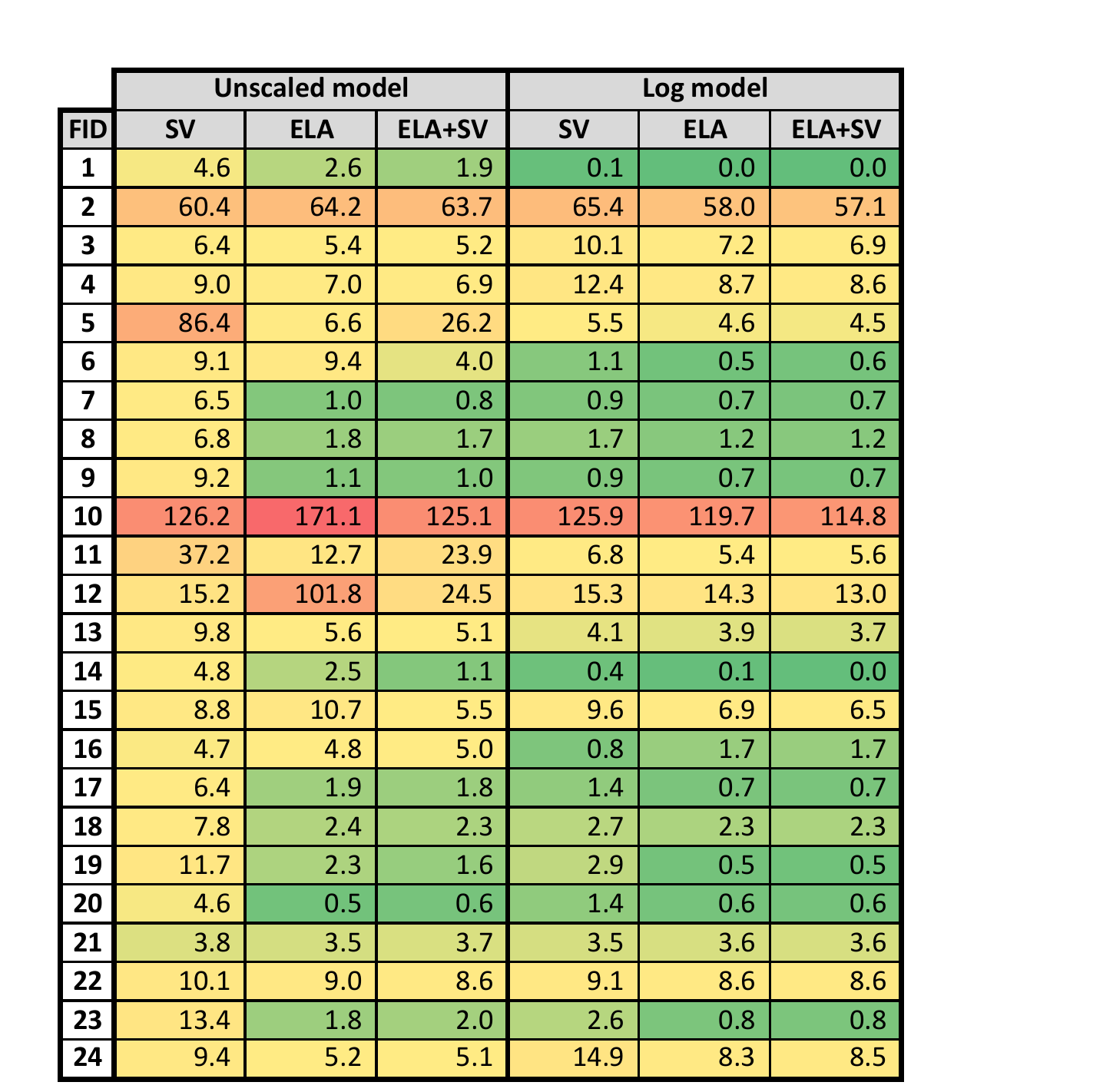}
\caption{Absolute prediction errors for both regression models aggregated per BBOB function in 3 different scenarios depending on the feature set used. The \emph{SV} column stands for the CMA-ES state variables, the \emph{ELA} for the landscape features, and the third one is the combination of both.} \label{fig:traj_pred}
\end{figure*}

Figure~\ref{fig:traj_pred} highlights the absolute prediction errors per BBOB function using two regression models, the unscaled and the log- one, when trained with 3 different feature sets: using only the trajectory landscape data, only the CMA-ES state variable data, and the combination of the two.  For the majority of the functions, using the combination of the trajectory data and the state variable data seems to help in improving the performance prediction accuracy, compared to the scenarios which use only one of those two feature sets.  

We also confirm that the log-model is indeed better at predicted fine-grained target precision (e.g., in the case of F1 (sphere function) or F6 (linear slope function), we know that those functions do not require many function evaluations to converge to the global optimum, and their recorded target precision values are already quite small as they are very near the optimal solution).  On the other hand, the unscaled model performs better where the target precision values are higher (e.g., for the functions such as F3, F15 (two versions of Rastrigin function), and also F24 (Lunacek bi-Rastigin), which are all highly multimodal, the number of function evaluations in our budget was not nearly enough to allow for finding a true optimum).

We also notice that using only the state variables for the unscaled model does not suffice for an accurate prediction in the most cases.  The reverse situation is nevertheless also possible: we see that for F12, using only the state variables yields the best accuracy in the unscaled model.  Furthermore, there are also exceptions where using only the landscape data results in a higher accuracy than using the combined features (e.g., F11 for both models, F5 for the unscaled model). 

\section{Comparison with Global Feature Values}
\label{sec:compglobal}

We then proceeded to compare the differences in the prediction accuracy from the sets described in the Section~\ref{sec:supervisedml} with the prediction accuracy using the global feature data, both alone and combined with the same CMA-ES state variable data as above.  To be able to perform a fair comparison, for the trajectory data we selected from the 20 executed CMA-ES runs those runs with the median target precision value per problem instance and their corresponding features and re-trained the unscaled and the log-model.  Global features-wise, both models were also trained using features computed from 2000 and 250 globally uniformly sampled points (the median value of 50 independent feature computations) for each function and instance.

\begin{figure*}[t]
\centering 
    \includegraphics[trim= 1.5cm 5cm 2cm 1cm, clip, width=\textwidth]{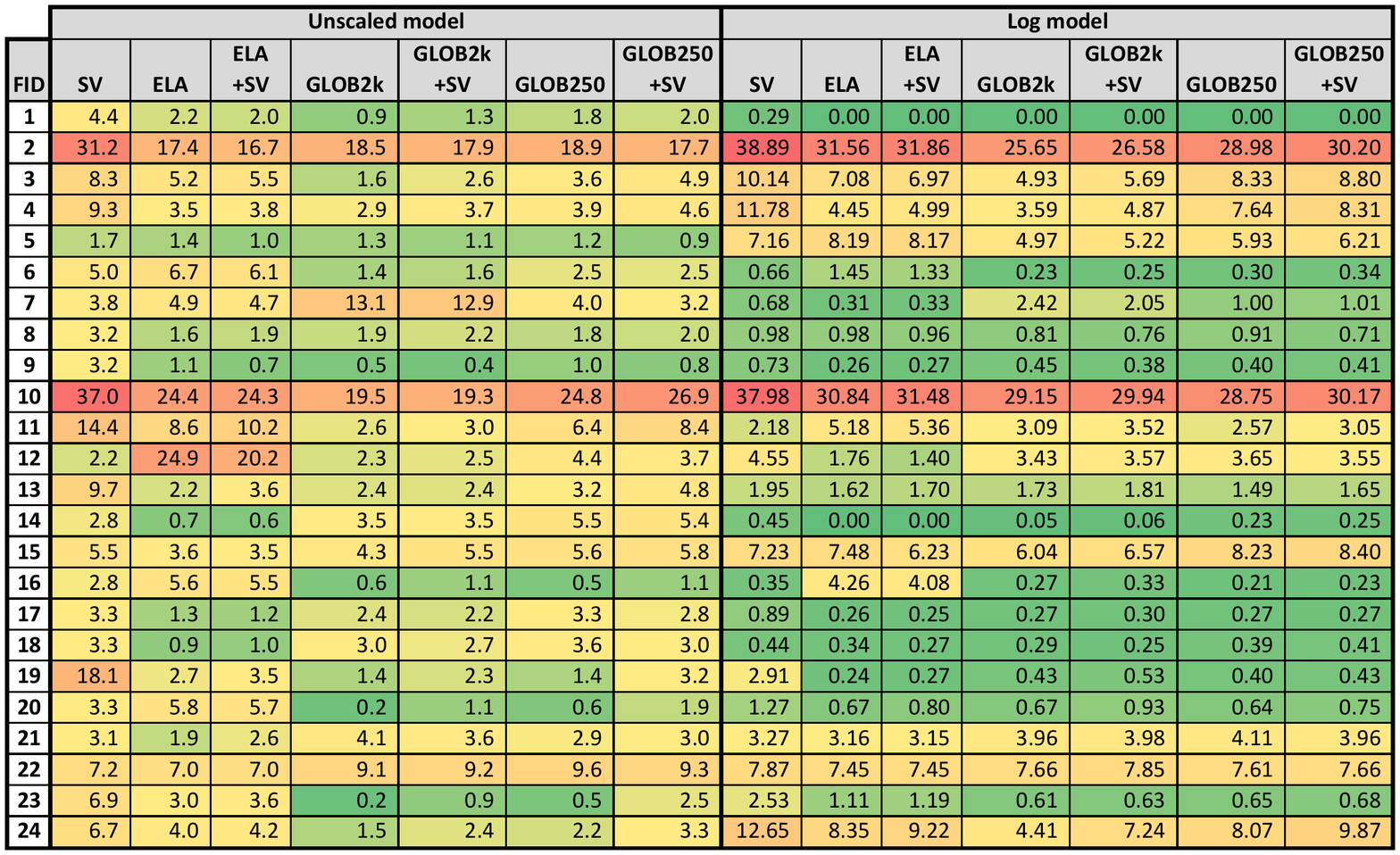}
\caption{Absolute prediction errors for both regression models for the median trajectory-based prediction (the first 3 columns of each block) and the median global feature prediction (the middle two columns of each block represent the errors when using the 2000-sample features, and the last two columns correspond to using the 250-sample features).} \label{fig:glob_pred}
\end{figure*}

Figure~\ref{fig:glob_pred} shows the absolute errors in prediction when the trajectory-based approach is compared with the results using the global features.  The highest accuracy is reported in cases when only the global landscape features were used, across almost all problems, with 2000-sample features yielding the best results.  Here, we do not observe a huge improvement when combining the global landscape features with the state variable data.  It seems that the number of samples used to compute the features can be crucial in reducing the errors in prediction, as global sampling could be linked to a potential higher discriminative power of features thus computed. Again, for certain functions such as F2 and F10 (both of which are different variants of the ellipsoidal function), we observe an overall low accuracy.

\section{Sensitivity Analyses}
\label{sec:sensitivity}

\textbf{Feature Selection.} 
To provide a sensitivity analysis based on the features used for the performance regression, we performed feature selection in the scenario of transfer learning, i.e., between different supervised tasks, where the features selected for the problem classification task have been evaluated on the performance regression task.

To do this, we have explored four state-of-the-art feature selection techniques: 
\textit{Boruta}~\cite{kursa2010boruta} is a feature selection and ranking algorithm based on random forests algorithm, which only selects features that are statistically significant. 
\textit{Recursive feature elimination (rfe)}~\cite{granitto2006recursive} learns a model assessing different sets of features by recursively eliminating features per loop until a good model is learnt. It requires an ML algorithm for evaluation, and here we use a random forest. 
\textit{Stepwise forward and backward selection (swfb)}~\cite{derksen1992backward} tries to fit the best regression model by iteratively selecting and removing features. In our experiments, we used it in both directions simultaneously. 
\textit{Correlation analysis with different threshold values (cor)}~\cite{benesty2009pearson} is based on the correlation analysis done only using the features (i.e., excluding the target). The result is a feature set where highly correlated features are omitted. In our case, we tested three different correlation thresholds: 0.50, 0.75, and 0.90. 
Note that while the first three feature selection methods require a supervised ML task, the last one is completely unsupervised and does not depend on the target. 

Our experimental design has been done using stratified 5-fold cross-validation. For a fair feature selection, we used the aforementioned methods on each training fold separately, then selected the intersection of the features returned by each training fold in the end. These features are further evaluated in the performance regression task.

\begin{table*}[t]
\centering
{\scriptsize{
\begin{tabular}{@{}lccccccc@{}}
\toprule
  & boruta & swfb & rfe & cor0.5 & cor0.75 & cor0.9 \\ \midrule
   \# selected ELA  features                   & 37     & 1    & 7   & 4      & 9       & 15     \\ \midrule
        \# selected state variable      features                   & 2      & 0    & 0   & 3      & 3       & 5      \\ \bottomrule 
\end{tabular}
}}
\caption{Number of ELA and state variable features for each selected feature portfolio. Details are available in Table~\ref{tab:selectedfeaturesdetails}.}
\label{tab:selectedfeatures}
\end{table*}

Table~\ref{tab:selectedfeatures} summarizes how many features were selected per portfolio, from the whole set of 38 ELA landscape features and 5 CMA-ES state variable features. 

\begin{table*}[t]
\centering
\resizebox{\textwidth}{!}{
\begin{tabular}{@{}c|cc|ccc:cccc:cccccc@{}}
\toprule
               &  &  & \textbf{SV}    & \textbf{ELA}   & \textbf{ELA} & \textbf{GLOB2k} & \textbf{GLOB2k} & \textbf{GLOB250} & \textbf{GLOB250} & \textbf{boruta} & \textbf{cor} & \textbf{cor} & \textbf{cor} & \textbf{rfe}   & \textbf{swfb}  \\ 
                              & &  & &   & \textbf{+SV} &  & \textbf{+SV} &  & \textbf{+SV} &  & \textbf{0.5}  & \textbf{0.75} & \textbf{0.9} &  & \\ \midrule
        &&& \multicolumn{13}{c}{\textbf{Best threshold $\tau$}}      \\                
\textbf{FID}            &  \textbf{min\_tp }      &  \textbf{ max\_tp}  &   \textbf{1.336} & \textbf{3.99}  & \textbf{4.742}   & \textbf{14.497}        & \textbf{9.46}              & \textbf{0.694}        & \textbf{2.605}            & \textbf{3.63}   & \textbf{1.813}  & \textbf{4.901}   & \textbf{1.717}  & \textbf{7.388} & \textbf{20}    \\  \midrule
1              & 0       & 0       & 0.43  & 0     & 0       & 0             & 0                 & 0            & 0                & 0      & 0      & 0       & 0      & 0     & 0.21  \\
2              & 9.25    & 87.47   & 44.69 & 25.47 & 24.46   & 24.49         & 23.65             & 28.96        & 27.98            & 25.16  & 36.78  & 35.28   & 34.62  & 24.51 & 53.85 \\
3              & 10.34   & 14.63   & 9.73  & 6.62  & 7.02    & 5.62          & 6.25              & 4.43         & 9.39             & 6.73   & 8.08   & 7.7     & 5.63   & 7.95  & 8.82  \\
4              & 10.29   & 14.53   & 11.86 & 5.08  & 5.08    & 4.81          & 5.73              & 5.1          & 7.49             & 5.18   & 9.26   & 6.71    & 5.59   & 7.53  & 8.41  \\
5              & 8.53    & 11.34   & 4.59  & 8.24  & 8.15    & 6.02          & 5.52              & 1.48         & 1.06             & 8.21   & 2.36   & 6.76    & 4.91   & 4.52  & 7.74  \\
6              & 0.04    & 0.11    & 0.78  & 3.87  & 2.15    & 0.24          & 0.27              & 0.33         & 0.37             & 4.46   & 1.02   & 1.19    & 1.65   & 0.89  & 1.43  \\
7              & 0.13    & 1.83    & 3.84  & 0.36  & 0.39    & 4.49          & 4.07              & 4.62         & 1.13             & 0.42   & 0.86   & 0.6     & 0.65   & 0.67  & 7.94  \\
8              & 1.22    & 2.59    & 3.02  & 1.06  & 1.1     & 1.25          & 1.22              & 2.03         & 0.84             & 0.95   & 2.02   & 0.93    & 1.04   & 1.42  & 4.88  \\
9              & 0.68    & 1.36    & 3.05  & 0.34  & 0.34    & 0.59          & 0.53              & 0.77         & 0.48             & 0.33   & 0.52   & 0.26    & 0.29   & 0.86  & 3.47  \\
10             & 8.44    & 84.69   & 43.33 & 32.06 & 32.85   & 21.39         & 23.16             & 29.29        & 33.35            & 32.13  & 42.18  & 32.73   & 35.01  & 29.64 & 47.68 \\
11             & 4.97    & 8.83    & 17.48 & 10.37 & 12.63   & 3.46          & 2.99              & 8.26         & 10.18            & 11.17  & 25.11  & 21.61   & 20.86  & 2.43  & 4.77  \\
12             & 2.53    & 6.32    & 4.76  & 19.34 & 15.89   & 4.03          & 4.16              & 5.76         & 4.13             & 19.45  & 4.38   & 2.31    & 9.69   & 19.32 & 3.56  \\
13             & 1.09    & 5.92    & 13.62 & 2.07  & 2.13    & 2.29          & 2.32              & 3.51         & 2.21             & 2.11   & 2.45   & 2.04    & 5.37   & 2.03  & 3.5   \\
14             & 0       & 0       & 1.65  & 0.01  & 0.01    & 0.07          & 0.08              & 1.89         & 0.37             & 0.01   & 0.61   & 0       & 0      & 0     & 5.18  \\
15             & 8.49    & 12.93   & 7.44  & 6.79  & 6.66    & 6.08          & 6.81              & 6.24         & 8.12             & 5.77   & 7.07   & 6.11    & 5.88   & 6.73  & 9.12  \\
16             & 0.18    & 0.83    & 0.87  & 6.31  & 5.85    & 0.32          & 0.35              & 0.25         & 0.25             & 6.45   & 0.48   & 4.42    & 4.77   & 2.43  & 1.21  \\
17             & 0.03    & 0.64    & 2.63  & 0.36  & 0.33    & 0.32          & 0.33              & 0.28         & 0.31             & 0.38   & 4.25   & 0.41    & 0.38   & 0.27  & 2.75  \\
18             & 0.16    & 0.69    & 1.54  & 0.34  & 0.34    & 0.33          & 0.3               & 4.96         & 0.45             & 0.34   & 0.54   & 0.28    & 0.3    & 0.54  & 1.24  \\
19             & 0.75    & 1.19    & 18.75 & 0.34  & 0.36    & 0.65          & 0.67              & 1.66         & 0.49             & 0.38   & 9.95   & 0.59    & 0.5    & 0.46  & 8.54  \\
20             & 1.7     & 1.79    & 3.77  & 0.76  & 0.84    & 0.99          & 1.17              & 0.99         & 0.93             & 0.69   & 3.2    & 1.22    & 3.16   & 1.77  & 5.16  \\
21             & 0       & 8.12    & 4.86  & 4.55  & 4.52    & 5.02          & 5.02              & 5.2          & 4.95             & 4.53   & 4.72   & 4.84    & 4.83   & 4.83  & 3.72  \\
22             & 0       & 25.48   & 12.04 & 11.91 & 11.91   & 11.9          & 11.91             & 12.88        & 11.94            & 11.91  & 11.89  & 11.93   & 11.7   & 11.92 & 5.34  \\
23             & 1.99    & 2.35    & 7.49  & 3.67  & 3.61    & 0.69          & 0.77              & 0.64         & 0.91             & 3.72   & 7.89   & 2.73    & 4.53   & 2.46  & 3.04  \\
24             & 15.73   & 20.73   & 10.06 & 4.76  & 4.94    & 5.15          & 6.52              & 2.95         & 4.29             & 4.81   & 11.13  & 9.49    & 7.22   & 7.43  & 11.65 \\ \midrule
\textbf{Overall RMSE, combined}             &    &   	 & \textbf{15.05} 	 & \textbf{10.41} 	 & \textbf{10.25} 	& \textbf{7.74} 	 & \textbf{7.92} 	& \textbf{9.48} 	 & \textbf{10.05} 	 & \textbf{10.43} 	& \textbf{13.66} 	 & \textbf{11.67} 	 & \textbf{11.86}   & \textbf{9.77} 	 & \textbf{15.73}  \\
Overall RMSE, unscaled             &    &    	& 15.08 	& 11.18 	&  10.88 	& 9.21 	 & 9.30 	 & 9.58 	& 10.19 	& 11.16 	&  14.11 	& 11.80 	&  12.00 & 10.93 	& 17.03  \\
Overall RMSE, log             &    &      	& 15.63 	& 13.05 	& 13.21 	 & 11.46 	& 11.88 	& 12.05 	& 12.87 & 13.14 	& 14.65 	& 13.89 	& 14.29	& 12.61 	& 15.73 
 \\\bottomrule
\end{tabular}}
\caption{RMSE values of the combined selector in three scenarios: when the prediction is based on the search trajectory landscape features and state variables (first 3 columns), on global features (next 4 columns), and finally on selected feature portfolios (last 6 columns).}
\label{tab:RMSEpermodel}
\end{table*} 

\textbf{Combined Selector Model and Sensitivity Analysis.} 

As common in ML, we measure the regression accuracy in terms of \emph{Root Mean Squared Error} (RMSE). Table~\ref{tab:RMSEpermodel} summarizes the RMSE values for the different feature portfolios when using (1) the unscaled model, (2) the scaled model, and (3) a combination of unscaled and the log-model (see last three rows of Table~\ref{tab:RMSEpermodel}). The threshold $\tau$ at which the predictive model changes is optimized for each feature portfolio individually, the obtained thresholds are summarized at the top of Table~\ref{tab:RMSEpermodel}. That is, 
we select the prediction of the log-model when the predicted precision (according to the log-model) is smaller than the threshold value $\tau$, and we use the prediction of the unscaled model otherwise. Note that the optimal threshold value $\tau$ varies significantly between the different feature portfolios. 

When comparing all the different portfolios (initial trajectory-based, global and selected trajectory-based ones), the good performance of the global feature sets is not surprising. Differences from the initial trajectory-based predictions are marginal for sets such as \emph{boruta}, \emph{cor0.75} and \emph{cor0.9}, whereas \emph{swbf} and \emph{cor0.5} perform constantly worse than \emph{ELA+SV}. Using the \emph{rfe} set, on the other hand, led to better results than using the original feature set.  \emph{SV} alone does not achieve good accuracy, but its contribution to ELA-only feature portfolio is around 3\% at the best threshold for the combined model, which is $\tau=4.901$. 
The absolute errors per instance are plotted in Figure~\ref{fig:abs_error_details}. 

\begin{figure}[t]
    \centering 
    \includegraphics[trim= 3cm 4cm 8cm 3cm, clip, width=\textwidth]{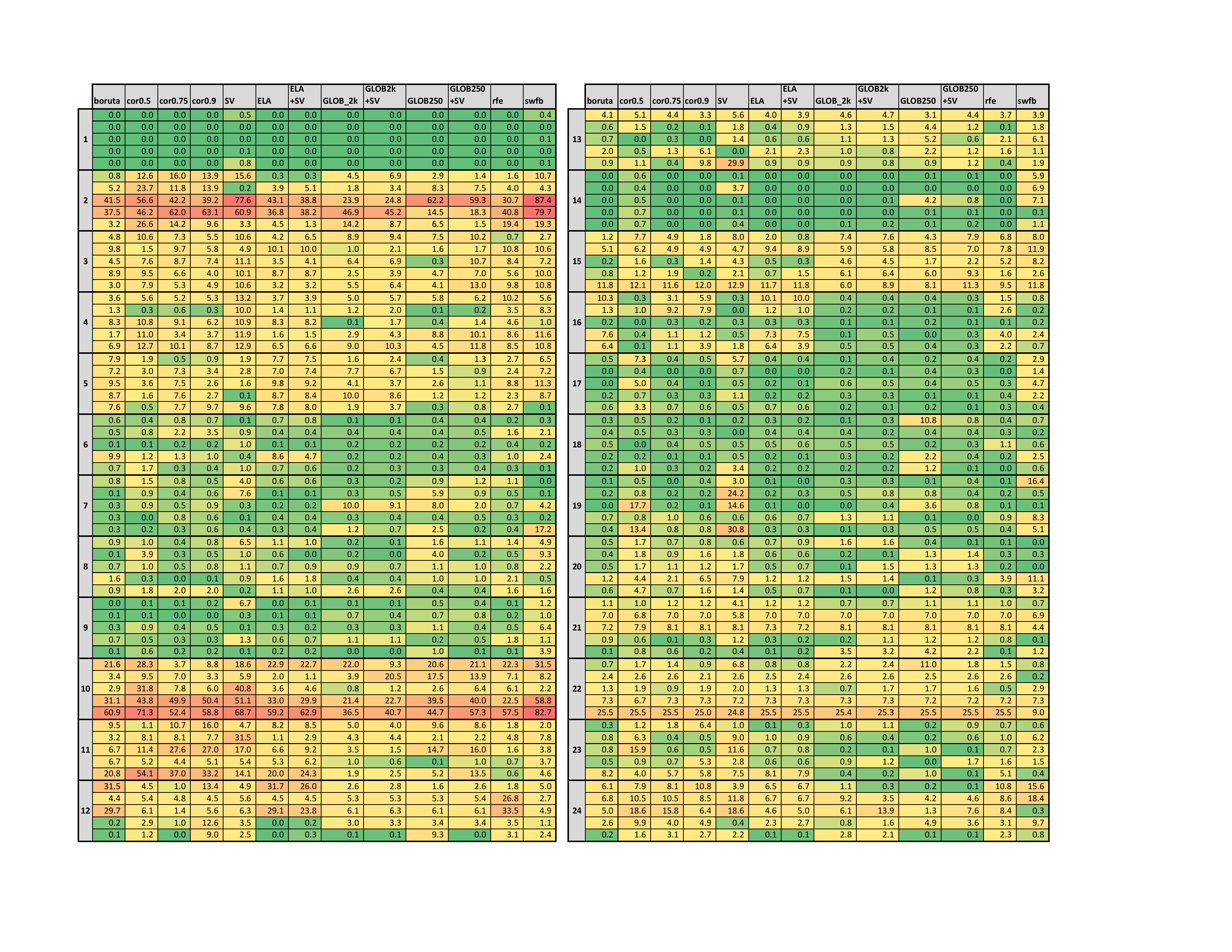}
    \caption{Absolute prediction errors of the combined models using portfolio-specific optimal thresholds $\tau$.}
    \label{fig:abs_error_details}
\end{figure}

\section{Conclusions and Future Work}
\label{sec:conclusions}

We analyzed in this paper the accuracy of predicting the CMA-ES solution quality after given budget based on the features computed from the samples on the CMA-ES search trajectory using two complementary regression models, the unscaled and the log-model. Adding information obtained from the CMA-ES internal state variables does not improve the prediction accuracy drastically compared to the trajectory-based data only.  Those results were then contrasted to the regression using the global features, where using the latter ones, especially those computed using a higher number of samples, yielded a consistently better accuracy.

Next, we tested whether we would achieve further gains in accuracy through feature selection. Although the overall results are comparable to the ones from initial trajectory-based portfolios, several selected feature sets resulted in worse accuracy than in the initial approach.  We ultimately pointed out the advantages of using our combined selector model over relying separately on predictions of the standalone unscaled or log-model across all different feature portfolios in all 3 scenarios.

In terms of \textbf{future work,} we plan on continuing this research by considering the following questions and tasks: \\
(0) Performance prediction of \textbf{other solvers}: How accurately can we use trajectory-based features of one algorithm to predict the performance of another algorithm?  In this work, we have only tried to predict performance for the same algorithm from whose trajectory the feature values have been computed. A next step would be to test if models for configuring the same algorithm can be trained. When this is successful, transfer learning from one algorithm to another one can be considered.   \\
(1) How can we more efficiently capture the \textbf{temporal component}, i.e., the information which sample was evaluated \emph{when} during the search? Using such longitudinal data, both in terms of extracted feature values and in terms of state variable evolution could possibly be done using recurrent neural networks~\cite{zhao2019learning}.\\
(2) \textbf{Combining global and trajectory-based sampling:} In our work, we only considered the case in which \emph{either} global sampling \emph{or} trajectory-based sampling is used. The accuracy of the models based on global sampling was better than that of the trajectory-based features. Even if we keep in mind that this comparison was unfair in that we provided the global feature values ``for free'', the results nevertheless suggest that a combination of global and trajectory-based feature computations could be worthwhile to investigate. How we can optimally balance the budget between global sampling, trajectory-based sampling, and remaining optimization budget is a challenging question in this context. \\
(3) \textbf{Warm-starting} the CMA-ES such that it starts the optimization process with the covariance matrix and other parameters that are extrapolated from the (uniformly or otherwise) distributed global samples might significantly improve the overall accuracy, as the CMA-ES will have a better overview of the whole problem instance "from the get-go". A similar approach has been suggested in~\cite{MohammadiRT15BOCMAES} when switching from a Bayesian optimization algorithm to CMA-ES. \\
(4) \textbf{Feature selection and ranking:} Instead of using transfer learning for feature selection between two different supervised ML tasks, feature selection within the same supervised task has not been considered in this paper. We also plan on making better use of variable importance estimations provided by feature ranking algorithms such as those based on ensemble of predictive clustering trees~\cite{petkovic2020feature} and those based on ReliefF and RReliefF~\cite{robnik2003theoretical}. \\
(5) \textbf{Feature design:} The work~\cite{DerbelLVAT19} suggests several algorithm-specific features for the SOO tree algorithm~\cite{munos2011SOOtree}. Such specific features can much more explicitly capture the characteristics of the algorithm-problem instance interaction. It could be worthwhile to study whether, possibly in addition to the longitudinal data mentioned in (1), such specific features can be identified for other common solvers, such as the CMA-ES. \\
(6) \textbf{Feature portfolio:} We note that our work above is based on the features available in the \emph{flacco} package~\cite{flacco}. Since the design of \emph{flacco}, however, several new feature sets have been suggested. Another straightforward way to extend our analyses would be in the inclusion of these feature sets, with the hope to improve the overall regression accuracy. In this respect, we find in particular the Search Trajectory Networks suggested in~\cite{OchoaM020continupusLON} worth investigating. \\
(7) \textbf{Representation learning of landscapes:} The  feature data will be additionally explored by applying representation learning methods that automatically learn new data representations by reducing the dimension of the data, automatically detecting correlations, and removing bias and redundancies presented in the feature data. The work presented in~\cite{eftimov2020linear} showed that linear matrix factorization representations of the ELA features values significantly detects better correlation between different problem instances.\\
(8) \textbf{Hyperparameter tuning of regression models: } Last, but not least, we are planning to explore algorithm portfolio that consists of different regression methods in order to find the most suitable one, together with finding its best hyperparameters for achieving better performance. In this study, we have used random forest for regression without tuning its parameters, since we have been interested in the contribution of different feature portfolios.   

\vspace{1ex}
\textbf{Acknowledgments.} 
This research benefited from the support of the Paris Ile-de-France region and of a public grant as part of the
Investissement d'avenir project, reference ANR-11-LABX-0056-LMH,
LabEx LMH. This work was also supported by projects from the Slovenian Research Agency: research core funding No. P2-0098 and project No. Z2-1867. We also acknowledge support by COST Action CA15140 ``Improving Applicability of Nature-Inspired Optimisation by Joining Theory and Practice (ImAppNIO)''. 

}
\newcommand{\etalchar}[1]{$^{#1}$}
\providecommand{\bysame}{\leavevmode\hbox to3em{\hrulefill}\thinspace}
\providecommand{\MR}{\relax\ifhmode\unskip\space\fi MR }
\providecommand{\MRhref}[2]{%
  \href{http://www.ams.org/mathscinet-getitem?mr=#1}{#2}
}
\providecommand{\href}[2]{#2}

\begin{table}
\scriptsize{
\centering
\resizebox{0.85\textwidth}{!}{
\begin{tabular}{@{}c|c|ccccccc@{}}
\toprule
 Feature& \# sets & boruta & swfb & rfe & cor0.5 & cor0.75 & cor0.9 \\ \midrule
ela\_distr.skewness                     & 4                  & x      & x    & x   &        &         & x      \\
ela\_distr.kurtosis                     & 4                  & x      &      & x   & x      & x       &        \\
ela\_distr.number\_of\_peaks            & 4                  & x      &      &     & x      & x       & x      \\
ela\_meta.lin\_simple.adj\_r2           & 2                  & x      &      &     &        &         & x      \\
ela\_meta.lin\_simple.intercept         & 1                  & x      &      &     &        &         &        \\
ela\_meta.lin\_simple.coef.min          & 2                  & x      &      &     &        &         & x      \\
ela\_meta.lin\_simple.coef.max          & 2                  & x      &      & x   &        &         &        \\
ela\_meta.lin\_simple.coef.max\_by\_min & 3                  &        &      &     & x      & x       & x      \\
ela\_meta.lin\_w\_interact.adj\_r2      & 1                  & x      &      &     &        &         &        \\
ela\_meta.quad\_simple.adj\_r2          & 2                  & x      &      & x   &        &         &        \\
ela\_meta.quad\_simple.cond             & 4                  & x      &      &     & x      & x       & x      \\
ela\_meta.quad\_w\_interact.adj\_r2     & 3                  & x      &      & x   &        &         & x      \\
disp.ratio\_mean\_02                    & 1                  & x      &      &     &        &         &        \\
disp.ratio\_mean\_05                    & 1                  & x      &      &     &        &         &        \\
disp.ratio\_mean\_10                    & 1                  & x      &      &     &        &         &        \\
disp.ratio\_mean\_25                    & 1                  & x      &      &     &        &         &        \\
disp.ratio\_median\_02                  & 2                  & x      &      &     &        &         & x      \\
disp.ratio\_median\_05                  & 1                  & x      &      &     &        &         &        \\
disp.ratio\_median\_10                  & 1                  & x      &      &     &        &         &        \\
disp.ratio\_median\_25                  & 1                  & x      &      &     &        &         &        \\
disp.diff\_mean\_02                     & 1                  & x      &      &     &        &         &        \\
disp.diff\_mean\_05                     & 1                  & x      &      &     &        &         &        \\
disp.diff\_mean\_10                     & 1                  & x      &      &     &        &         &        \\
disp.diff\_mean\_25                     & 1                  & x      &      &     &        &         &        \\
disp.diff\_median\_02                   & 3                  & x      &      &     &        & x       & x      \\
disp.diff\_median\_05                   & 1                  & x      &      &     &        &         &        \\
disp.diff\_median\_10                   & 1                  & x      &      &     &        &         &        \\
disp.diff\_median\_25                   & 1                  & x      &      &     &        &         &        \\
nbc.nn\_nb.sd\_ratio                    & 2                  & x      &      &     &        &         & x      \\
nbc.nn\_nb.mean\_ratio                  & 1                  & x      &      &     &        &         &        \\
nbc.nn\_nb.cor                          & 2                  & x      &      &     &        &         & x      \\
nbc.dist\_ratio.coeff\_var              & 3                  & x      &      &     &        & x       & x      \\
nbc.nb\_fitness.cor                     & 2                  & x      &      & x   &        &         &        \\
ic.h.max                                & 3                  & x      &      &     &        & x       & x      \\
ic.eps.s                                & 1                  & x      &      &     &        &         &        \\
ic.eps.max                              & 1                  & x      &      &     &        &         &        \\
ic.eps.ratio                            & 4                  & x      &      & x   &        & x       & x      \\
ic.m0                                   & 3                  & x      &      &     &        & x       & x      \\
\midrule
step\_size                              & 4                  & x      &      &     & x      & x       & x      \\
mahalanobis\_dist                       & 2                  &        &      &     & x      &         & x      \\
c\_evol\_path                           & 3                  &        &      &     & x      & x       & x      \\
sigma\_evol\_path                       & 2                  &        &      &     &        & x       & x      \\
cma\_simil\_lh                          & 2                  & x      &      &     &        &         & x      \\ 
\midrule
\end{tabular}}
\caption{Feature portfolios.
}
\label{tab:selectedfeaturesdetails}
}
\end{table}

\end{document}